\documentclass{article}

\usepackage{pifont}
\newcommand{\cmark}{\ding{51}}%
\newcommand{\xmark}{\ding{55}}%
\usepackage{siunitx}
\usepackage{subfig}
\usepackage{graphicx}
\usepackage{makecell}
\usepackage[colorlinks,linkcolor=blue]{hyperref}

\usepackage[utf8]{inputenc} % allow utf-8 input
\usepackage[T1]{fontenc}    % use 8-bit T1 fonts
\usepackage{booktabs}       % professional-quality tables
\usepackage{amsfonts, amsmath, amssymb}       % blackboard math symbols
\usepackage{nicefrac}       % compact symbols for 1/2, etc.
\usepackage{microtype}      % microtypography
\usepackage{hyperref}
\usepackage{array,multirow}
 \usepackage{float}

\usepackage[toc,page,header]{appendix}
\usepackage{minitoc}
\usepackage{comment}

\PassOptionsToPackage{numbers, compress}{natbib}

\usepackage[preprint]{neurips_2021}

\usepackage{graphicx}
\usepackage{bm}

\usepackage[linesnumbered, ruled]{algorithm2e}

\usepackage{multicol}
\usepackage{mathtools}

\usepackage{color}
\usepackage{pifont}

\SetCommentSty{mycommfont}

\usepackage{xcolor}
\definecolor{dark-red}{rgb}{0.4,0.15,0.15}
\definecolor{dark-blue}{rgb}{0.15,0.15,0.4}
\definecolor{medium-blue}{rgb}{0,0,0.5}
\hypersetup{
   colorlinks, linkcolor={dark-blue},
   citecolor={dark-blue}, urlcolor={medium-blue}
}

\title{Self-supervised Interpretable Representation Learning from Videos}
\title{Zero-Shot Segmentation by Motion Decoupling from Videos}
\title{Foreground Segmentation by Appearance-Motion Decomposition from Videos}
\title{Zero-Shot Segmentation by Appearance-Motion Decomposition from Videos}
\title{Zero-Shot Object Segmentation by Appearance-Motion Decomposition from Videos}

\title{Zero-Shot Object Segmentation from \\ Unsupervised Segment-Motion Decomposition}
\title{Zero-Shot Object Segmentation from \\ 
Unsupervised Segment-Motion  Factorization}
\title{Zero-Shot Object Segmentation by Appearance-Motion Decomposition}
\title{The Emergence of Objectness: \\ Learning Zero-Shot Segmentation from Videos}

\author{%
\setlength{\tabcolsep}{18pt}
\begin{tabular}{@{}cccc@{}}
    Runtao Liu$^{1,2}$\thanks{Equal contribution. Work done when Runtao was a StarBridge intern at MSRA.} & 
    Zhirong Wu$^{1}$\footnotemark[1] &
    Stella X. Yu$^{3}$ & 
    Stephen Lin$^{1}$\\
\end{tabular}\\[10pt]
\setlength{\tabcolsep}{10pt}
\begin{tabular}{@{}ccc@{}}
    Microsoft Research Asia$^1$ & 
    John Hopkins University$^2$ &
    UC Berkeley / ICSI$^3$\\
\end{tabular}\\
\setlength{\tabcolsep}{6pt}
\begin{tabular}{@{\hspace{-10pt}}ccc@{}}
    \texttt{runtao219@gmail.com}& 
    \texttt{stellayu@berkeley.edu}& 
    \texttt{\{wuzhiron,stevelin\}@microsoft.com}
\end{tabular}    
}

\begin{document}

\maketitle

\begin{abstract}

Humans can easily segment moving objects without knowing what they are.  
That objectness could emerge from continuous visual observations motivates us to model grouping and movement concurrently from {\it unlabeled} videos.  
Our premise is that a video has different views of the same scene related by moving components, and the right region segmentation and region flow would allow mutual view synthesis which can be checked from the data itself without any external supervision.

Our model starts with two separate pathways: an appearance pathway that outputs feature-based region segmentation for a single image, and a motion pathway that outputs motion features for a pair of images.  
It then binds them in a conjoint representation called {\em segment flow} that pools flow offsets over each region and provides a gross characterization of moving regions for the entire scene.  
By training the model to minimize view synthesis errors based on segment flow, our appearance and motion pathways learn region segmentation and flow estimation automatically without building them up from low-level edges or optical flows respectively.

Our model demonstrates the surprising emergence of objectness in the appearance pathway, surpassing prior works on zero-shot object segmentation from an image, moving object segmentation from a video with unsupervised test-time adaptation, and semantic image segmentation by supervised fine-tuning.  Our work is the first truly end-to-end zero-shot object segmentation from videos. It not only develops generic objectness for segmentation and tracking, but also outperforms prevalent image-based contrastive learning methods without augmentation engineering.

\end{abstract}

\def\imw#1#2{\includegraphics[width=#2\linewidth]{#1}}
\def\imh#1#2{\includegraphics[height=#2\textheight]{#1}}
\def\imwh#1#2#3{\includegraphics[width=#2\linewidth,height=#3\textheight]{#1}}
\newcommand{\tb}[3]{\setlength{\tabcolsep}{#2mm}\begin{tabular}{#1}#3\end{tabular}}

\def\figTeaserOld{
\begin{figure}[t]
\centering
\subfloat{\small
    \tb{@{}c@{}}{0}{
    \imw{figs/fig_teaser.pdf}{0.99}
    }
}
\caption{Motivation of the approach. Segmenting moving objects based on dense optical flow is prone to noise, non-smoothness, and articulated motion. We propose a segment flow representation which decouples appearance and motion. As a result, object segments can be segmented without full liability to dense optical flows.}
\label{fig:teaser}
\end{figure}
}

\def\figTeaser#1{
\begin{figure}[#1]
\centering
\imw{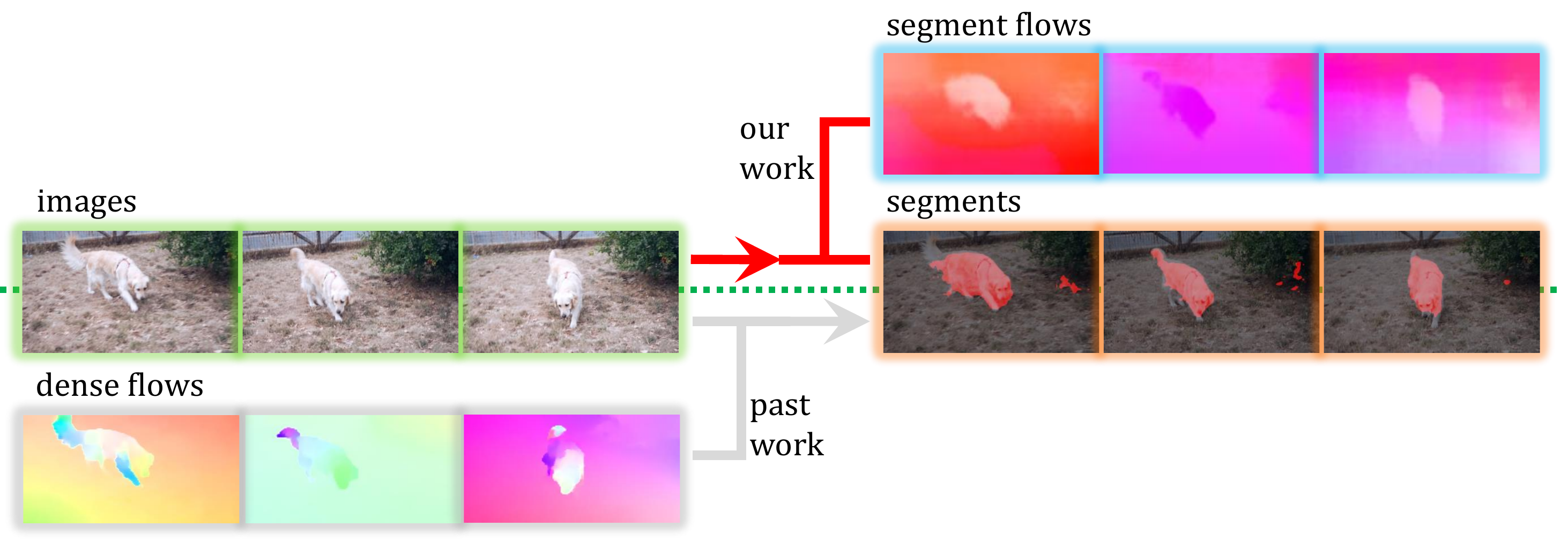}{1}
\caption{
Our zero-shot object segmentation is learned from an unsupervised factorization of images into segments and their motions, whereas past work segments objects based on dense pixel-wise optical flows, which are brittle in the presence of noise, articulated movement, and abrupt motion. }
\vspace{-15pt}
\label{fig:teaser}
\end{figure}
}

\section{Introduction}

In recent years, contrastive learning~\cite{wu2018unsupervised,he2020momentum,chen2020simple} has become a powerful model for obtaining high-level representations on generic images~\cite{goyal2021self}.
Despite their encouraging performance, contrastive models have two critical limitations. 
First, they heavily rely on hand-crafted image augmentations~\cite{dosovitskiy2015discriminative} to induce invariance, yet fail to account for complex variations such as object deformations and 3D viewpoints. 
Second, they still require additional labeled data and a fine-tuning stage for downstream applications, preventing use in a standalone fashion.

In this paper, we seek a zero-shot image model for detecting and segmenting objects by learning from unlabeled videos with minimal %image flipping 
augmentation. As opposed to static images, a dynamic sequence of observations provides information about what is moving in the scene and how it moves. Such patterns not only reveal the boundary segment of an object, 
%its 3D structure, 
but also indicate hierarchical part organizations and even object semantics. Thus, as an easily accessible source of unlabeled data, videos provide rich natural supervision for learning image representations.

A popular objective for self-supervised learning from videos is view synthesis. Concretely, given a source frame, a function is learned to warp the source frame to the target frame using photometric consistency as supervision. Dense optical flow~\cite{liu2020learning} can be self-supervised in this manner. One could also learn a monocular depth network from videos with additional camera parameters~\cite{godard2019digging}. 
The key idea is to find an appropriate representation which not only parameterizes the warping function, but also transfers to the target task. For example, the multi-plane image representation~\cite{zhou2018stereo} is proposed to extrapolate between stereo pairs using view synthesis.

\figTeaser{tp}

Unlike prior works that use view synthesis for low-level vision tasks, our goal is to tackle object segmentation which involves mid-level and high-level visual recognition.
To this end, dense optical flow field which represents the low-level correspondence in a local manner would not suffice (see Figure~\ref{fig:teaser}).
We therefore seek a new representation which could capture a gross characterization of moving regions for the entire scene.
Deriving a representation explicitly for moving regions would allow the model to localize and segment objects.

Our approach decomposes view synthesis into two visual pathways: an appearance pathway to model ``what is moving'' by segmenting a static RGB image into separate regions, and a motion pathway to model ``how it moves'' by extracting motion features on a pair of images.
The motion features are then used to predict flow offsets for individual regions assuming common fate~\cite{wertheimer1923untersuchungen} for all pixels within a region.
The segment masks as well as their flow vectors jointly reconstruct a new representation called {\em segment flow}, which is used for view synthesis. 
In this way, object appearance and motion are decoupled, such that the appearance model for predicting segmentation would benefit from rich RGB signals. By conditioning on a region, the motion pathway is also tasked to solve a much simpler problem than dense flow.
The two pathways are jointly trained with a reconstruction loss.

After self-supervised pretraining, we find that generic objectness detection and object segmentation automatically emerge from the model. Our model has the versatility for a variety of applications. First, the appearance pathway can be directly applied to novel images for dominant object segmentation in a zero-shot fashion. Second, it can also be fine-tuned for semantic segmentation on a small labeled dataset. Finally, with unsupervised test-time adaptation, the overall model can be transferred to novel videos for moving object segmentation without labels. Experimentally, we demonstrate strong performance on all of these applications, showing considerable improvements against the baselines.

The contributions of this work can be summarized as follows: 1) the first truly end-to-end zero-shot object segmentation from unlabeled videos; 2) a conceptually novel segment flow representation which goes beyond traditional dense optical flow; 3) a versatile model that can be applied to various image and video segmentation tasks.
Our code is available at  \href{https://github.com/rt219/The-Emergence-of-Objectness}{https://github.com/rt219/The-Emergence-of-Objectness}.

\section{Related Works}

\textbf{Video object segmentation. }
Segmentation of moving objects requires finding correspondences along the time dimension. A dominant line of work focuses on learning a representation for temporally propagating segmentation masks. Such a representation may be learned with pixel-level object masks in videos with long-term relations~\cite{oh2019video,zhang2020transductive}, or learned through self-supervision such as colorization~\cite{vondrick2018tracking} and cycle-consistency~\cite{jabri2020space}. Given the annotation of object masks in the initial frame, the model tracks the object and propagates the segmentation through the remaining frames.

Fully unsupervised video object segmentation, without initial frame annotations, has received relatively little attention. NLC~\cite{faktor2014video} and ARP~\cite{koh2017primary} take a temporal clustering approach to this problem. Though they do not require segmentation annotations, elements of these algorithms depend on edge and saliency labels, and thus are not completely unsupervised.
FTS~\cite{papazoglou2013fast} calculates the segmentation by obtaining a motion boundary from the optical flow map between frames. SAGE~\cite{wang2015saliency} takes into account multiple cues of edges, motion segmentation, and image saliency for video object segmentation.
Contextual information separation~\cite{yang2019unsupervised} segments moving objects by exploiting the motion independence between the foreground and the background. A concurrent work based on motion grouping~\cite{yang2021self} clusters pixels with similar motion vectors. Both of these works rely on an off-the-shelf optical flow representation, which may be trained with~\cite{sun2018pwc,teed2020raft} or without~\cite{liu2020learning} supervision.

\textbf{Motion Segmentation.}
Classical methods for motion segmentation~\cite{shi1998motion,kumar2008learning,sun2012layered} cluster distinctive motion regions from the background based on two-frame optical flow.
Supervised learning approaches~\cite{tokmakov2017learning,tokmakov2019learning} map the optical flow field to segmentation masks.
The requirement of dense and accurate optical flow may be problematic when the flow vectors are not smooth over time and  vulnerable to articulated objects with inhomogeneous motion~\cite{ochs2013segmentation}.
we turn our attention to modeling appearance on RGB representations, which provide rich cues (e.g. texture, color and edges) for perceptual organization, alleviating the need for dense pixel correspondence.

\textbf{Motion Trajectory Segmentation.}
Moving object segmentation has been shown to be effective when motion is considered over a large time interval~\cite{ochs2013segmentation}. An approach based on trajectory clustering~\cite{keuper2015motion} builds point trajectories over hundreds of frames, extracts descriptors for the point trajectories, and clusters them to obtain segmentation results. Though promising, such a global approach is computationally demanding.

\textbf{Layered representations.}
A simple linear model~\cite{wang1993layered,wang1994representing} can factorize a video into layers of foreground objects and background, assuming independence among the objects and background. This layered representation was used to derive better optical flow estimates~\cite{sun2013fully,sun2012layered,sun2010layered} and also for view-interpolation and time retargeting applications~\cite{brostow1999motion,alayrac2019visual,zitnick2004high,lu2020layered}.
Different from prior works, our work demonstrates the emergence of objectness through such layered representations.

\textbf{Unsupervised learning for segmentation.} 
Human annotation of pixel-level segmentation is not only time-consuming, but also often inaccurate along object boundaries. Learning segmentation without labels is thus of great interest in practice. Segsort~\cite{hwang2019segsort} predicts segmentation by learning to group super-pixels of similar appearance and context from static images. Later work~\cite{van2021unsupervised}  contrasts holistic mask proposals obtained from traditional bottom-up grouping.

A related line of work focuses on learning part segmentation from images and videos of the same object category, such as humans and faces.  SCOPS~\cite{hung2019scops} is a representative method learned in a self-supervised fashion. The general idea follows unsupervised landmark detection~\cite{jakab2018unsupervised}, where geometric invariance, representation equivariance and perceptual reconstructions are considered. Co-part segmentation~\cite{xu2019unsupervised} is also explored in videos, where motion provides a strong cue for part organization. A motion-supervised approach~\cite{siarohin2020motion} models part motion between adjacent frames via affine parameters. 
Another work~\cite{sabour2020unsupervised} implements a similar idea in capsule networks.
Our work differs significantly in studying learning from generic videos instead of from a single visual category.

\textbf{Image representation learning using motions.} 
Motion contains rich cues about object location, shape, and part hierarchy. Motion segmentation has been used as a self-supervision signal for learning image-level object representations~\cite{pathak2017learning}. Motion propagation~\cite{zhan2019self} predicts a dense optical flow field from sparse optical flow vectors, conditioned on an RGB image. Our work also produces an image representation from unlabeled videos. Unlike prior works, our image representation is a by-product of our full framework for video understanding. 

\def\imw#1#2{\includegraphics[width=#2\linewidth]{#1}}
\def\imh#1#2{\includegraphics[height=#2\textheight]{#1}}
\def\imwh#1#2#3{\includegraphics[width=#2\linewidth,height=#3\textheight]{#1}}

\def\figModelOld{
\begin{figure}[t]
\centering
\subfloat{\small
    \tb{@{}c@{}}{0}{
    \imw{figs/pipeline2.pdf}{0.99}
    }
}
\caption{Overview of the model training procedure. Our model consists of two independent networks: an upper branch that predicts segmentation masks on individual RGB images, and a lower branch that extracts motion features from a pair of RGB images.
The motion features are pooled for each segmentation mask, and the pooled motion feature is used to predict a single flow vector for the segment following the common fate assumption. The flow vectors are subsequently disseminated in each segment to recover a segment flow map. The learning of the segment flow map is supervised by a reconstruction objective between the two input frames.  }
\label{fig:arch}
\end{figure}
}

\iffalse
\def\figModel#1{
\begin{figure}[#1]
\centering
\imw{figs/videoSeg_v2.pdf}{1}
\caption{
We learn a single-image segmentation network and a dual-frame motion network with an unsupervised image reconstruction loss.  We sample two frames, $i$ and $j$, from a video.  Frame $i$ goes through the \textcolor{red}{segmentation} network and outputs a set of masks, whereas frames $i$ and $j$ go through the \textcolor{cyan}{motion} network and output a feature map.  The feature is pooled per mask and a flow is predicted.  All the segments and their flows are combined into a segment flow representation from frame $i\to j$, which are used to \textcolor{cyan}{warp} frame $i$ into $j$, and \textcolor{orange}{compared} against frame $j$ to train the two networks.
}
\label{fig:model}
\end{figure}
}
\fi

\section{Segmentation by Appearance-Motion Decomposition}

\begin{figure}[tp]
\centering
\imw{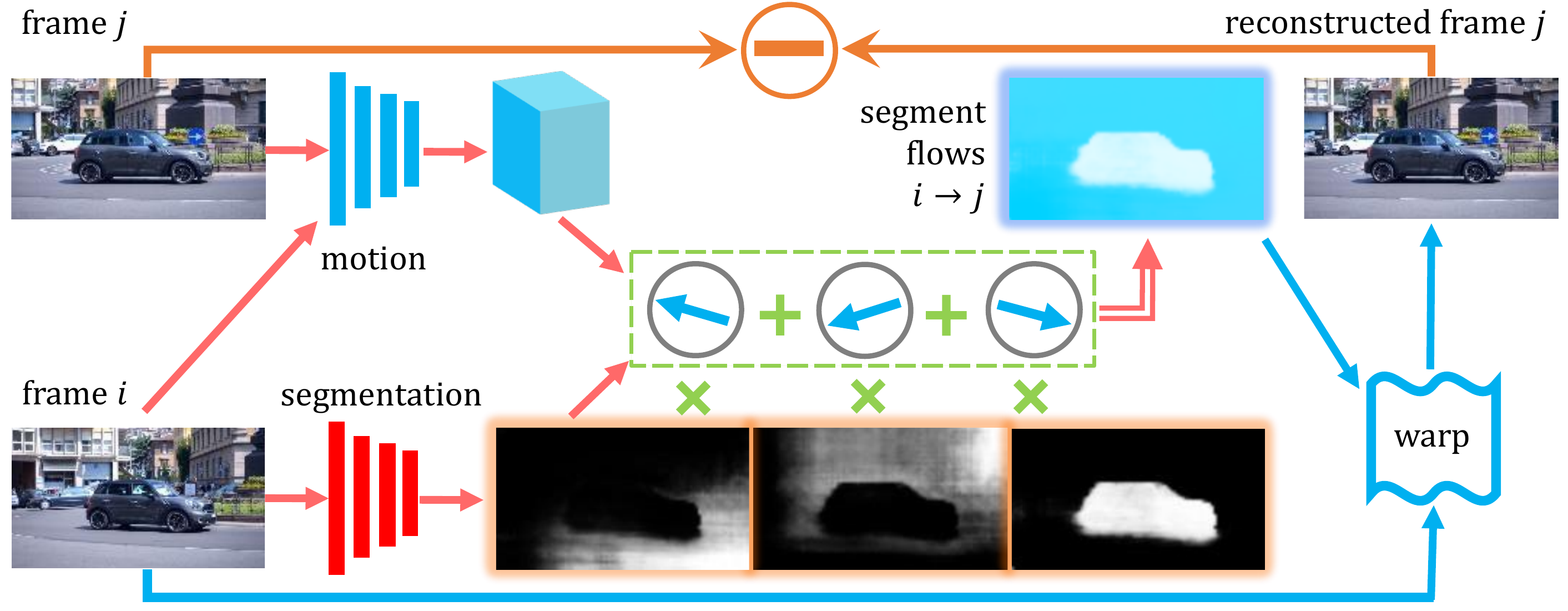}{1}
\caption{
We learn a single-image segmentation network and a dual-frame motion network with an unsupervised image reconstruction loss.  We sample two frames, $i$ and $j$, from a video.  Frame $i$ goes through the \textcolor{red}{segmentation} network and outputs a set of masks, whereas frames $i$ and $j$ go through the \textcolor{cyan}{motion} network and output a feature map.  The feature is pooled per mask and a flow is predicted.  All the segments and their flows are combined into a segment flow representation from frame $i\to j$, which are used to \textcolor{cyan}{warp} frame $i$ into $j$, and \textcolor{orange}{compared} against frame $j$ to train the two networks.
}
\vspace{-5pt}
\label{fig:model}
\end{figure}

The goal of this work is to learn a zero-shot model to detect and segment objects by merely exposing it to unlabeled videos.
We are only interested in detecting the objectness in this paper instead of further categorizing the objects into specific classes. 
We assume that a single moving object appears in one video. When multiple moving objects occur, the model needs to group these objects as one.

We take a learning-based approach to this problem. During training, we are given a collection of unlabeled videos for self-supervised learning.
The pretrained model should be directly applicable for inference on a novel image or a video to produce the object segmentation masks. 
The overall pipeline for training our model is illustrated in Figure~\ref{fig:model}. 
Our approach appearance-motion decomposition (AMD) takes a pair of RGB frames $X_i$ and $X_j$ sampled from a video for learning. 
The model consists of an appearance pathway $f_A(X_i)$ and a motion pathway $f_M(X_i,X_j)$. 
The two pathways jointly construct a segment flow representation $F$, which is used to warp frame $X_i$ into $X_j$. The overall model is self-supervised by the reconstruction objective on the frame $X_j$.
In the following, we describe the details for each module in our model.

\subsection{Appearance Pathway for Segmentation}

The appearance pathway is a fully convolutional neural network for segmenting a static RGB image into separate regions.
Formally, given the image $X_i \in \mathbb{R}^{3\times h \times w}$, it is segmented into $c$ regions by
\begin{equation}
    S = f_A(X_i) \in \mathbb{R}^{c\times h \times w}.
\end{equation}
In practice, the mask $S$ is a soft probability distribution normalized across $c$ channels. 
$c$ is an important hyper-parameter of our approach. A large $c$ may lead to over-segmentation, and a small $c$ may not locate the object. Empirically, we use a default value of $c=5$, and this is examined later in an ablation study.

We note that our segmentation network is designed to operate on static images and thus the network can be transferred to downstream image-based vision tasks. 
In Section~\ref{sec:saliency}, we demonstrate that the pretrained segmentation network can be used to detect salient objects in a zero-shot fashion. Fine-tuning the appearance pathway on a labeled dataset is examined in Section~\ref{sec:semantic}.

\subsection{Motion Pathway for Correspondence}

The purpose of the motion pathway is to extract pixel-wise motion features between a pair of images in order to predict the region flow vector detailed in the next subsection.
We follow the network architecture of PWC-Net~\cite{sun2018pwc} for predicting dense optical flow, where the feature for each pixel describes the perceptual similarity to its spatial neighbors in the other frame.
Formally, given input frames $X_i$ and $X_j$, the network extracts  features $V$ by
\begin{align}
V = f_M(X_i, X_j) \in \mathbb{R}^{d_v \times h \times w}
\label{eq:fm}
\end{align}
where $d_v$ is the dimension of motion features. 

\subsection{Segment Flow Representation}

Given the decoupled appearance pathway and motion pathway, the segment and its motion can be binded for view synthesis.
Concretely, we pool the pixel-wise motion features within each segmentation mask to obtain the mask motion feature as a single vector,
\begin{align}
 V_m = \frac{\sum(V \odot S_m)}{\sum S_m} \in \mathbb{R}^{d_v}, \quad m=1,...,c
\label{eq:pool}
\end{align}
where the summation operation is taken across the spatial coordinates, and $m$ is used to index the segmentation masks. The optical flow vector for each segmentation mask is read out from the motion feature by
\begin{align}
 F_m = g(V_m) \in \mathbb{R}^2, \quad m=1,...,c
\label{eq:flow}
\end{align}
where the head network $g(\cdot)$ is chosen as a two-layered multilayer perceptron (MLP).

So far, we decompose a pair of images $X_i \to X_j$ into a set of segmentation masks $S_m$ and their associated flow vectors $F_m$. 
This decomposition is based on the assumption that pixels within a mask share the same motion, a condition that simplifies optical flow estimation. 
This assumption may not hold for articulated objects and inhomogeneous motion.
However, it becomes less problematic when all views in a video are taken for optimization, with
the appearance pathway able to aggregate a smoothly moving region into a meaningful segment.

We reconstruct a novel flow representation for the full image by composing the layers of segments with their motion vectors,
\begin{align}
F = \sum_{m} F_m \odot S_m, \quad m=1,...,c,
\label{eq:G}
\end{align}
where $\odot$ denotes the outer product. Since the flow representation $F$ is segment-based, we refer to it as {\em segment flow}.
This decoupled representation allows each component to cross-supervise each other. 
Given an optical flow offset, the segmentation network could be supervised to find pixels that share this offset.
Given a segmentation mask, the correspondence network could be supervised to find the flow offset for this mask.

This approach for supervising object segmentation using motion information is fundamentally different from motion segmentation methods.
Our segmentation mask is predicted from a static appearance model that does not require dense and accurate flow for supervision. It utilizes flow at the region level, which can be approximated from sparse and noisy pixel-level estimates.

\subsection{Reconstruction Objective}

With the segment flow offset map, we are able to warp frame $X_i$ to $X_j$ by
\begin{align}
 \hat{X_j}(p)=X_i(p+F(p)),
\end{align}
where $p$ is a spatial location index. The ground-truth frame $X_j$ provides supervision for reconstructed frame $\hat{X_j}$ through the following objective,
\begin{align}
\mathcal{L}=D(X_j, \hat{X_j}),
\label{eq:loss}
\end{align}
where $D$ is a metric defining distance between two images. Among the numerous choices for $D$, such as photometric losses~\cite{wang2004image}, deep-feature-based losses~\cite{johnson2016perceptual}\cite{zhang2018unreasonable}, and contrastive losses~\cite{park2020contrastive}, we adopt the pixel-wise photometric loss of SSIM~\cite{wang2004image} in this work for simplicity.  

\subsection{Object Segment Selection}
Since our model outputs $c$ masks for each image, the mask corresponding to the object instead of the background needs to be determined. 
We have empirically observed that the primary moving objects all appear in a particular mask channel across the training videos. This channel can be heuristically identified as the one whose segmentation mask has the maximum averaged segment motion.
The object segment from this mask layer is used for evaluating zero-shot downstream tasks.

\def\imw#1#2{\includegraphics[width=#2\linewidth]{#1}}
\def\imh#1#2{\includegraphics[height=#2\textheight]{#1}}
\def\imwh#1#2#3{\includegraphics[width=#2\linewidth,height=#3\textheight]{#1}}

\section{Experiments}
\label{sec:exp}

We demonstrate that AMD model can be transferred to three downstream applications.
First,  the appearance pathway is directly applied on static images for salient object detection in a zero-shot fashion.
Second, both the appearance and the motion pathway are transferred to video object segmentation in novel videos with zero human labels.
Third, we fine-tune the appearance pathway on labeled data for semantic segmentation. 

\subsection{Training and Implementation Details}
\label{sec:imp}

AMD is pretrained on the large object-centric video dataset Youtube-VOS~\cite{xu2018youtube}.
The training split for Youtube-VOS contains about 4,000 videos covering 94 categories of objects. The total duration for the dataset is $334$ minutes.
We train the model on the data with a sampling rate 24 frames per second, without using the original segmentation labels.

We train all model parameters from scratch without external pretraining. For the segmentation network, we use ResNet50~\cite{he2016deep} as our backbone followed by a fully convolutional head containing two convolutional blocks. For the motion network, we adopt PWC-Net~\cite{sun2018pwc} architecture because of its effectiveness in estimating optical flows. 
We resize the short edge of the input image to $400$ pixels, and random crop a square image of size $384\times 384 $ with random horizontal flipping augmentation.
No other augmentations is engineered.
We adopt the symmetric loss that considers either frame as the target frame and sums the two reconstruction errors.
For training the overall model, we use the Adam optimizer with a learning rate of $\num{1e-4}$ and a weight decay of $\num{1e-6}$. 
We train AMD on eight V100 GPUs, with each processing two pairs of sampled adjacent frames.
The network is optimized for 400K iterations.

\subsection{Zero-Shot Saliency Detection}
\label{sec:saliency}

\def\prow#1#2#3#4{
\imw{figs/fig_saliency/#1.png}{0.244}&
\hspace{0.01cm}
\imw{figs/fig_saliency/#2.png}{0.244}&
\hspace{0.01cm}
\imw{figs/fig_saliency/#3.png}{0.244}&
\hspace{0.005cm}
\vrule
\hspace{0.005cm}
\imw{figs/fig_saliency/#4.png}{0.244}\\
}
\begin{figure}[t]
\centering
\subfloat{\small
    \tb{@{}cccc@{}}{0.05}{
    \prow{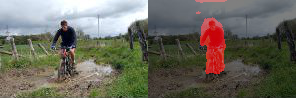}{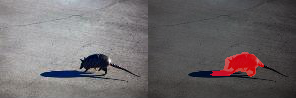}{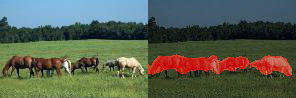}{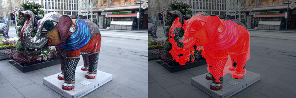}
    \prow{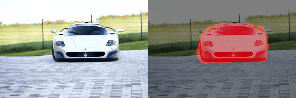}{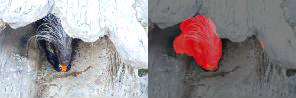}{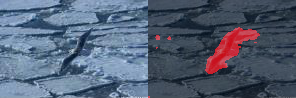}{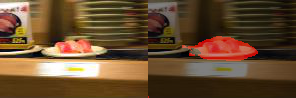}
    \prow{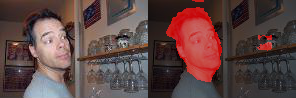}{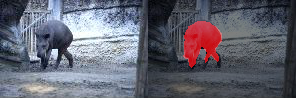}{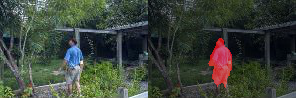}{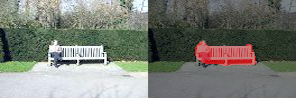}
    \prow{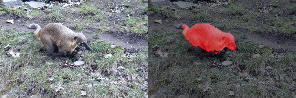}{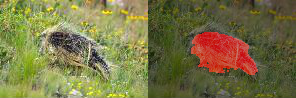}{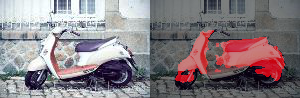}{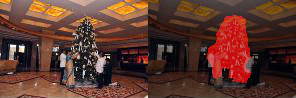}
     & movable objects & &
 stationary objects
    }
}
\caption{Qualitative salient object detection results. We directly transfer our pretrained segmentation network to novel images on the DUTS dataset without any finetuning. Surprisingly, we find that the model pretrained on videos to segment moving objects can generalize to detect stationary unmovable objects in a static image, e.g. the statue, the plate, the bench and the tree in the last column.}
\label{fig:saliency}
\vspace{-7pt}
\end{figure}

Once pretrained, AMD's appearance pathway can be directly transferred to object segmentation in novel stationary images without any downstream fine-tuning. To evaluate the quality of the segmentation, we benchmark the results on salient object detection benchmark.

The salient object detection performance is measured on the \textbf{DUTS~\cite{wang2017learning}} benchmark, which contains 5,019 test images with pixel-level ground truth annotations.
We follow two widely used metrics in this area: the $F_\beta$ score and the per-pixel mean squared errors (MAE). $F_\beta$ is defined as the weighted harmonic mean of the precision ($P$) and recall ($R$) scores: $F_\beta = \frac{(1+\beta^2) P \times R}{\beta^2 P + R}$, with $\beta^2 = 0.3$.
MAE is simply the per-pixel averaged error of the soft prediction scores.

\textbf{Experimental results.} 
We compare our saliency estimation results against several traditional methods based on low-level cues.
Useful low-level cues and priors include background priors~\cite{zhu2014saliency}, objectness~\cite{jiang2013saliency,jiang2013salient}, and color contrast~\cite{cheng2014global}.
As shown in Table~\ref{tab:saliency}, our method achieves an $F_\beta$ score $60.2$ and an MAE score of $0.13$, outperforming all traditional approaches by a notable margin.
We note that AMD is not designed specifically for this task nor for this particular dataset, and its strong performance demonstrates the generalization ability of the model.

In related work on unsupervised learning of saliency detection~\cite{zhang2018deep,zeng2019multi,nguyen2019deepusps}, the priors of traditional low-level methods are ensembled.
Though they do not use saliency annotations, their models are pretrained on ImageNet classification and even semantic segmentation with pixel-level annotations. These methods are thus not fully unsupervised, so they are omitted from the comparisons.

In Figure~\ref{fig:saliency}, we show some qualitative results on salient object detection.
Surprisingly, we find that our model pretrained on videos to segment moving objects not only detects movable objects in images, but also generalizes to detect stationary unmovable objects, such as statues, benches, trees and plates shown in the last column. This suggests that our model learns a generic objectness prior from the unlabeled videos.
We hypothesize that our model may learn objectness from the camera motion as well. Camera motion may cause the object and the background at various depth to have different observed projective 2D optical flow even though the objects are static.

\subsection{Zero-shot Video Object Segmentation}
\label{sec:vos}

\begin{table}[t]
	\begin{minipage}{0.4\linewidth}
		\centering
		\caption{Salient object detection performance on the DUTS dataset. Our model outperforms traditional low-level methods by notable margins.}
		\vspace{3pt}
		\setlength{\tabcolsep}{12pt}
		\begin{tabular}[t]{c|cc}
		\hline
		Model   & $F_{\beta}$  & MAE  \\
		\hline
		RBD\cite{zhu2014saliency}  & 51.0   & 0.20 \\
		HS\cite{zou2015harf}       & 52.1   & 0.23 \\
		MC\cite{jiang2013saliency} & 52.9   & 0.19 \\
		DSR\cite{li2013saliency} & 55.8 & 0.14 \\
		DRFI\cite{jiang2013salient} & 55.2 & 0.15 \\
		\hline
		\textbf{AMD}                       & \textbf{60.2}      & \textbf{0.13} \\
		\hline
		\end{tabular}
		\label{tab:saliency}
	\end{minipage}
	\quad
	\begin{minipage}{0.55\linewidth}  
		\centering
		\caption{Transfer performance for semantic segmentation on VOC2012. Our method outperforms TimeCycle and compares favorably with contrastive methods.}
		\vspace{3pt}
		\setlength{\tabcolsep}{10pt}
		\begin{tabular}[t]{c|c|c|c}
		\hline
		Model   & Data &  Aug. & mIoU   \\
		\hline
		Scratch & -- & -- &  48.0  \\
		TimeCyle\cite{wang2019learning} & VLOG & light & 52.8  \\
		%SimSiam\cite{chen2020exploring} & \xmark & 57.7  \\
		MoCo-v2\cite{he2020momentum}   & YTB  & light & 61.5 \\
%		\textbf{AMD}(100K) & YTB  &  light & 60.4 \\
		\textbf{AMD} & YTB  &  light & \textbf{62.0} \\
		\hline
		%SimSiam\cite{chen2020exploring} & \cmark & 59.4 \\
		MoCo-v2\cite{he2020momentum}       &  YTB  &  heavy & \textbf{62.8} \\
		\textbf{AMD} & YTB  &  heavy & 62.1 \\
		\hline
		MoCo-v2\cite{he2020momentum}       &  IMN  &  heavy & \textbf{72.4} \\
		\hline
		\end{tabular}
		\label{tab:seg}
	\end{minipage}
	\vspace{-10pt}
\end{table}

We transfer the pretrained AMD model to object segmentation on novel videos.
Since the segmentation prediction from our model is based on static images, inference on images sequentially in a video essentially estimates objectness.
In order to exploit motion information, we use a test-time adaptation approach.
Concretely, given a novel video, we optimize the training objective in Eq.~\ref{eq:loss} on pairs of frames sampled from the novel testing video. The adaptation takes 100 iterations per video. 

We evaluate zero-shot video object segmentation on three testing datasets. 
{\bf DAVIS 2016~\cite{Perazzi2016}} contains 20 validation videos with 1,376 annotated frames.  
{\bf SegTrackv2~\cite{li2013video}} contains 14 videos with 976 annotated frames. Following prior works, we combine multiple foreground objects in the annotation into a single object for evaluation. 
{\bf FBMS59 ~\cite{ochs2013segmentation}} contains 59 videos with 720 annotated frames. The dataset is relatively challenging because the object may be static for a period of time. We pre-process the ground truth labels following prior work~\cite{yang2019unsupervised}.
For evaluation, we report the Jaccard score, which is equivalent to the intersection over union (IoU) between the prediction and the ground truth segmentation.

\textbf{Experimental results.} 
We consider baseline methods claiming to be unsupervised for the full pipeline, including traditional non-learning-based approaches~\cite{wang2017saliency,faktor2014video,papazoglou2013fast,keuper2015motion,koh2017primary} and recent self-supervised learning methods~\cite{yang2019unsupervised,yang2021self}.
In Table~\ref{tab:FID}, we summarize the results for all the methods on the three datasets.
Among these methods, NLC~\cite{faktor2014video} actually relies on an edge model trained with human-annotated edge boundaries, and ARP~\cite{koh2017primary} depends on a segmentation model trained on a human-annotated saliency dataset. We thus gray their entries in the table.
For all the traditional methods, since their original papers do not report results on most of these benchmarks, we simply provide the performance values reported in the CIS paper~\cite{yang2019unsupervised}.

We evaluate the performance for AMD with and without test-time adaptation. No adaptation boils down to per-image saliency estimation using only the appearance pathway. Adaptation transfers both appearance and motion pathways.
On DAVIS 2016, our method achieves a Jaccard score of $57.8\%$, surpassing all traditional unsupervised models. 
For CIS~\cite{yang2019unsupervised}, their best performing model uses a significant amount of post-processing, including model ensembing, multi-crop, temporal smoothing and spatial smoothing.
We thus refer to their performance obtained from a single model without post-processing.
Our model is slightly worse than CIS on DAVIS, by $1.4\%$.
However, on SegTrackv2 and FBMS59, our method outperforms CIS by large margins of $11.4\%$ and $10.7\%$, respectively.
Motion grouping~\cite{yang2021self} is a work concurrent with ours.
It is essentially  a motion segmentation approach which relies on an off-the-shelf pre-computed dense optical flow model.
Motion grouping performs worse than our method on DAVIS2016 and SegTrackv2 when a low-performance unsupervised optical flow model ARFlow is used~\cite{liu2020learning}.
With a state-of-the-art supervised optical flow model~\cite{teed2020raft} that is trained on ground truth flow, their performance improves significantly.
Among all the discussed methods, ours is the first end-to-end self-supervised learning approach which does not require a pretrained optical flow model.

In Figure~\ref{fig:vos}, we show qualitative comparisons to the baseline CIS~\cite{yang2019unsupervised}. We display dense flow from pretrained PWC-Net, the CIS results, our segment flows, and our segmentation results. For most of these examples, our segment flow only coarsely reflects the true pixel-level optical flow. However, our segmentation results are significantly better and less noisy, as our model is relatively insensitive to optical flow quality.
In the first and the third examples, our model produces high-quality object segmentations even though the motion cue for the objects is very subtle.

\def\rowone#1#2#3#4#5#6#7{
\imw{figs/fig_vos/#7/#1.png}{0.152}&
\hspace{0.01cm}
\imw{figs/fig_vos/#7/#2.png}{0.152}&
\hspace{0.01cm}
\imw{figs/fig_vos/#7/#3.png}{0.152}&
\hspace{0.01cm}
\imw{figs/fig_vos/#7/#4.png}{0.152}&
\hspace{0.01cm}
\imw{figs/fig_vos/#7/#5.png}{0.152}&
\hspace{0.01cm}
\imw{figs/fig_vos/#7/#6.png}{0.152}\\
}
\begin{figure}[t]
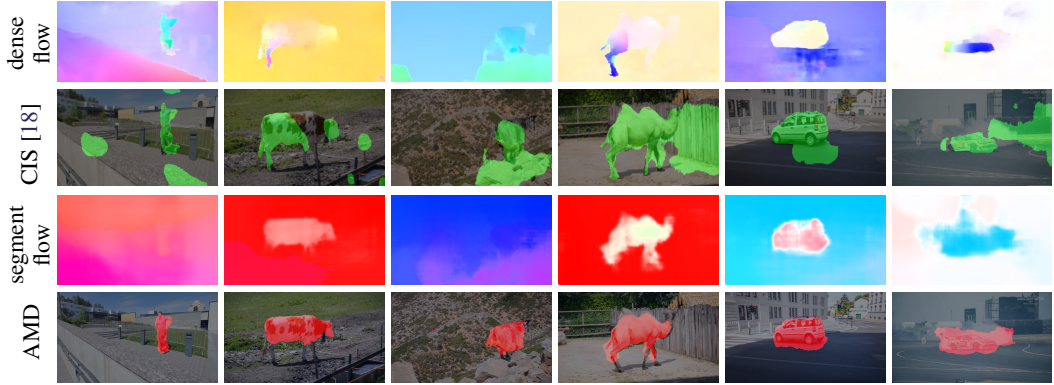

\centering
\subfloat{\small
    \tb{@{}ccccccc@{}}{0.03}{
    \rotatebox{90}{\makecell[c]{\hspace{0.15cm}dense\\\hspace{0.15cm}flow}} & \rowone{1}{5}{9}{17}{15}{18}{cis}
    \vspace{0.03cm}
    \rotatebox{90}{\hspace{0.0cm}CIS~\cite{yang2019unsupervised}}  & \rowone{0}{4}{8}{16}{14}{19}{cis} 
    %\hline
    \rotatebox{90}{\makecell[c]{segment\\flow}} & \rowone{10}{0}{8}{14}{16}{18}{ours}
    \rotatebox{90}{\hspace{0.3cm}AMD} & \rowone{11}{1}{9}{15}{17}{19}{ours}
    }
}
\caption{Qualitative comparisons to motion segmentation based method CIS~\cite{yang2019unsupervised} with its input dense flow and our segmentation results with segment flow representation. CIS is prone to noise, articulated motion, and camera motion in the dense flow estimations. %
By decomposing appearance from motion, our model AMD suffers less from these vulnerabilities of dense optical flow. This leads to results much better and more robust than the motion segmentation based approach.}
\label{fig:vos}
\vspace{-3pt}
\end{figure}

\begin{table}[t]
\centering
\caption{Performance evaluations for unsupervised video object segmentation on DAVIS 2016, SegTrackv2 and FBMS59 datasets. The numbers are measured in terms of Jaccard score. The table is split into traditional non-learning-based and recent self-supervised learning methods. The model results which rely on other kinds of human supervisions (Sup.) are \textcolor[gray]{0.5}{grayed}. Dependence for pretrained dense flow method is also listed for each model. MG's results on SegTrackv2 and FMBS59 using ARFlow are reproduced by ours and marked with $*$. 
We evaluate AMD with appearance pathway only and with both pathways for test time adaptation.
AMD performs favorably to the baseline CIS on DAVIS 2016, while showing large gains on the other two benchmarks.} %
\vspace{5pt}
\setlength{\tabcolsep}{6pt}
\begin{tabular}[t]{ccccc|ccc}
\hline
\hline
 & Model & e2e & Sup. & Flow & DAVIS 2016  & SegTrackv2 & FBMS59  \\
\hline
\parbox[t]{0.1mm}{\multirow{5}{*}{\rotatebox[origin=c]{90}{traditional}}} & SAGE\cite{wang2017saliency}   & \xmark  & \xmark     &  LDOF\cite{brox2010large} & 42.6 & 57.6 & 61.2 \\
& NLC\cite{faktor2014video}     & {\xmark}  & {edge}     &  {SIFTFlow}\cite{liu2009beyond} & \textcolor[gray]{0.5}{55.1}  & \textcolor[gray]{0.5}{67.2} & \textcolor[gray]{0.5}{51.5} \\
& CUT\cite{keuper2015motion}    & \xmark  & \xmark     &  LDOF\cite{brox2010large} & 55.2 & 54.3 & 57.2 \\
& FTS\cite{papazoglou2013fast}  & \xmark  & \xmark     &  LDOF\cite{sundaram2010dense} & 55.8 & 47.8 & 47.7 \\
& {ARP}\cite{koh2017primary}      & {\xmark}  & {saliency} &  {CPMFlow}\cite{hu2016cpm} & \textcolor[gray]{0.5}{76.2} & \textcolor[gray]{0.5}{57.2} & \textcolor[gray]{0.5}{59.8} \\
\hline
\parbox[t]{0.1mm}{\multirow{4}{*}{\rotatebox[origin=c]{90}{learning}}} & CIS\cite{yang2019unsupervised}  & \xmark  & \xmark & PWC\cite{sun2018pwc}    & 59.2 & 45.6 & 36.8 \\
& MG\cite{yang2021self}           & \xmark  & \xmark & ARFlow\cite{liu2020learning} & 53.2 & 37.8$^*$    &  50.4$^*$   \\
%\hline 
& \textbf{AMD} {\footnotesize (per-img)} & \cmark  & \xmark & \xmark  & 45.7  & 28.7 & 42.9 \\
& \textbf{AMD}  {\footnotesize (per-vid)} & \cmark  & \xmark & \xmark  & 57.8   & 57.0 & 47.5 \\
\hline
\hline
\end{tabular}
\label{tab:FID}
\vspace{-3pt}
\end{table}

\subsection{Semantic Segmentation}
\label{sec:semantic}

Given that our pretrained segmentation network can produce meaningful generic object segmentations, we further examine its semantic modeling ability on semantic segmentation. 
We conduct this experiment on the \textbf{Pascal VOC 2012~\cite{everingham2010pascal}} dataset.
The dataset contains 20 object categories with 10,582 training images and 1,449 validation images.
Given a pretrained model, we finetune the model on the training set and evaluate the performance on the validation set.
The finetuning takes 40,000 iterations with a batch size of 16 and an initial learning rate of 0.01. 
The learning rate undergoes polynomial decay with a power parameter of 0.9.

\textbf{Experimental results.}
We compare the pretrained model to a image-based contrastive model, MoCo-v2~\cite{he2020momentum} and a self-supervised video pretraining model, TimeCycle~\cite{wang2019learning}.
TimeCycle~\cite{wang2019learning} is pretrained on the VLOG dataset, which is larger than our Youtube-VOS dataset.
For MoCo-v2, we also pretrain the contrastive model on the Youtube-VOS dataset, to ablate the role of pretraining datasets.
Since the base version of our method does not utilize heavy augmentations as in contrastive models, we also study the effects of data augmentations.
The results are reported in Table~\ref{tab:seg}.
Our method outperforms the video pretraining approach TimeCylce significantly by $9.2\%$.
Compared with MoCo-v2, when light augmentation (resizing, cropping) is used, our model slightly outperforms MoCo-v2 by $0.5\%$. However, when heavy data augmentation (color jitter, grayscale, blurring) is applied, our method underperforms MoCo-v2 by $0.7\%$.
This is possibly because our model is non-contrastive in nature, and thus unable to take advantage of information effectively in augmentations.
MoCo-v2 performs much stronger when pretrained on ImageNet, possibly because the semantic distribution of ImageNet is well aligned with VOC2012.
Overall, our model outperforms a prior self-supervised video model TimeCycle and compares favorably with contrastive model MoCo-v2 under the same data.

\def\rowone#1#2#3#4#5#6{
\imw{figs/fig_segnum/#1}{0.16}&
\hspace{0.003cm}
\imw{figs/fig_segnum/#2}{0.16}&
\hspace{0.003cm}
\imw{figs/fig_segnum/#3}{0.16}&
\hspace{0.0015cm}
\vrule
\hspace{0.0015cm}
\imw{figs/fig_segnum/#4}{0.16}&
\hspace{0.003cm}
\imw{figs/fig_segnum/#5}{0.16}&
\hspace{0.003cm}
\imw{figs/fig_segnum/#6}{0.16}\\
}

\begin{figure}[t]
    \begin{minipage}{0.7\linewidth}  
    	\centering
    	\subfloat{\small
    		\tb{@{}cccccc@{}}{0.05}{
    		\rowone{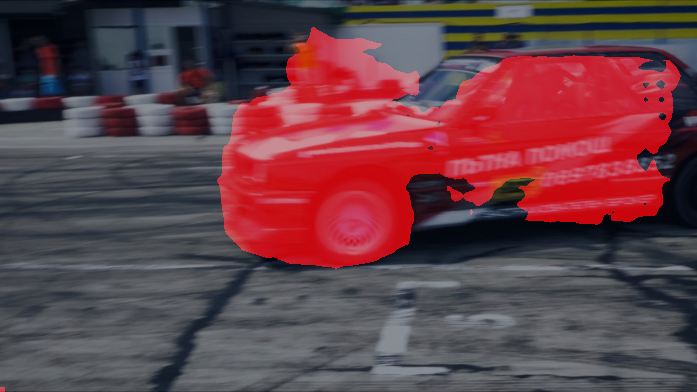}{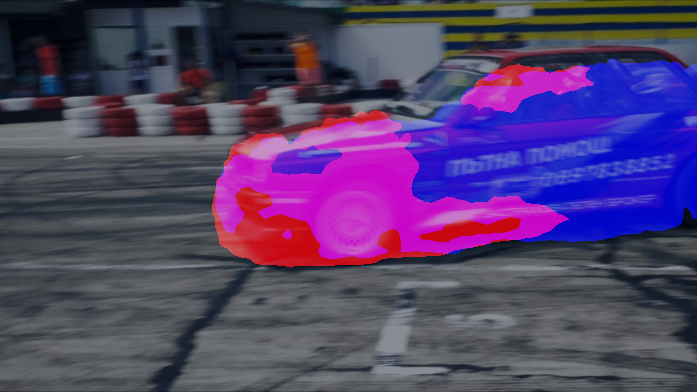}{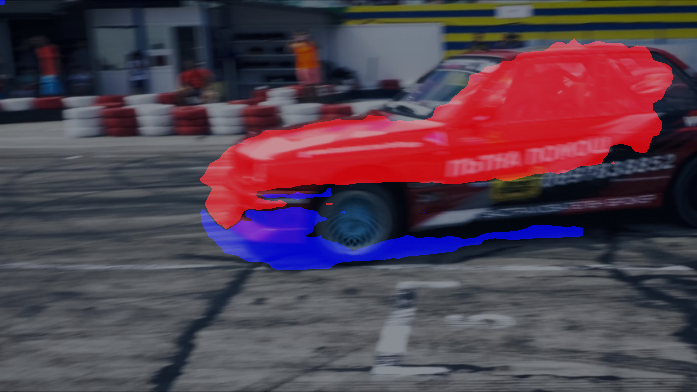}{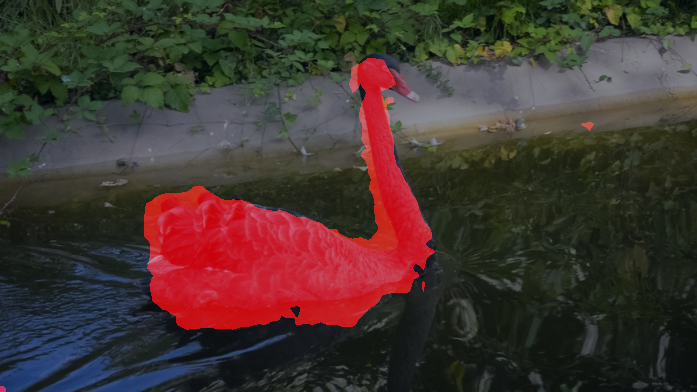}{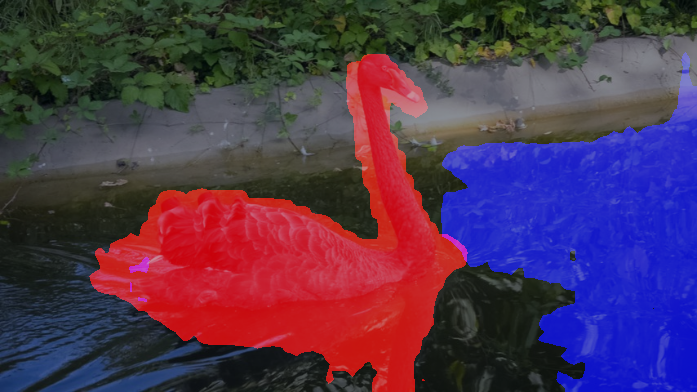}{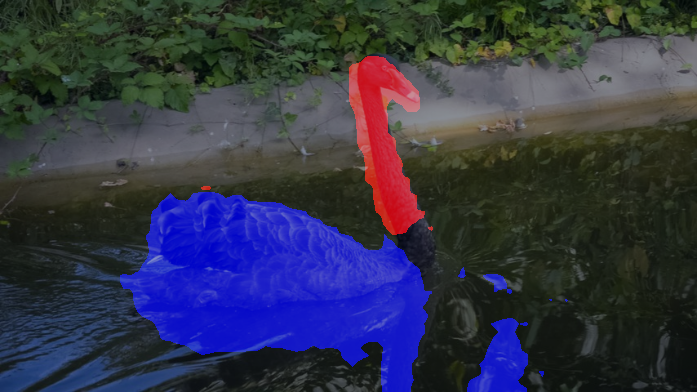}
    		\rowone{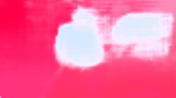}{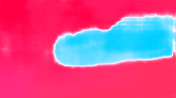}{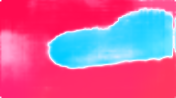}{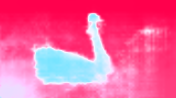}{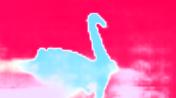}{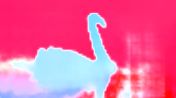} 
    		$c=5$ & $c=6$ & $c=8$ & $c=5$ & $c=6$ & $c=8$ \\
    		}
    	}
    	\label{fig:ab}
    \end{minipage}
    \quad
    \begin{minipage}{0.2\linewidth}
    	\centering
    	\setlength{\tabcolsep}{4pt}
    	\begin{tabular}[t]{c|c}
    	\hline
        \#segments & DAVIS ($J$)   \\
    	\hline
    	$c=5$     & \textbf{57.8}  \\
    	$c=6$     & 45.3           \\
    	$c=8$     & 41.0           \\
    	\hline
    	\end{tabular}
    	\label{tab:ab}
    \end{minipage}
    \captionof{figure}{Ablation study on different number of segments. Two examples with segmentation masks and segment flows are shown. The object region is split over multiple masks when $c$ becomes large. The over-segmentation decreases the performance of video object segmentation on DAVIS2016.}
    \vspace{-5pt}
    \label{fig:ab}
\end{figure}

\subsection{Ablation Study}
\label{sec:ablation}

The variable $c$, the number of segmentation channels, is an important hyper-parameter of our model. 
We vary the value of $c$ ($5,6,8$) for pretraining the model, and examine its transfer performance for video object segmentation on DAVIS2016.
In Figure~\ref{fig:ab}, we visualize the model predictions under different number of segments. We observe that a large $c$ tends to lead to over-segmentation, and a small $c$ tends to lead to large regions.
The car and the swan is split into multiple regions even if the motion for separated regions are very close.
The model trained with $c=5$ segments a full object, while the model trained with $c=8$ separates the object into parts.
When pretraining the model with $c<4$, the training becomes unstable.
Quantitatively, the video object segmentation performance on DAVIS2016 decreases as we increase the number of segments.

\begin{table}[t]

\end{table}

\section{Summary}
In this paper, we show that objectiveness could emerge from a computational model by exposing it to unlabeled videos. 
We present a novel model which decouples the appearance pathway and the motion pathway, and later binds them into a joint segment flow representation.
As opposed to prior works that rely heavily on accurate dense optical flows for predicting object segmentation, our method learns only from raw pixel observations. 
The motion representation in our model is a lot weaker, however the object segmentation is more robust.
The proposed model AMD is the first end-to-end learning approach for zero-shot object segmentation without using any pretrained modules.
Its performance is validated on a number of image  and video object segmentation benchmarks.

\bibliographystyle{unsrt}
\bibliography{neurips\_2021}

\newpage
\def\imw#1#2{\includegraphics[width=#2\linewidth]{#1}}
\def\imh#1#2{\includegraphics[height=#2\textheight]{#1}}
\def\imwh#1#2#3{\includegraphics[width=#2\linewidth,height=#3\textheight]{#1}}

\appendix

\section{Supplementary}

In this supplementary, we provide the network architecture details in Section A.1. In Section A.2, we present more qualitative video object segmentation results on 3 datasets, DAVIS 2016~\cite{Perazzi2016}, FBMS59~\cite{ochs2013segmentation} and SegTrackv2~\cite{li2013video}. We also present the per class quantitative results on these datasets. 

\subsection{Network Details}
The network details are shown in Table \ref{tab:detailseg} and Table \ref{tab:detailpwc}. Table \ref{tab:detailseg} shows the detailed network architecture for the segment prediction head of our segmentation network. Our correspondence network adopt the similar framework as PWCNet~\cite{sun2018pwc} which contains a feature extractor, a flow estimator and a context network. The feature extractor is the same as that of PWCNet while we don't use the context network in our correspondence network. Table \ref{tab:detailpwc} shows the detailed layers of the flow estimator. 

\subsection{More Results}

More video object segmentation results are shown in Figure~\ref{fig:suppsegt}, Figure~\ref{fig:suppdavis} and Figure~\ref{fig:suppfbms} for SegTrackv2, DAVIS 2016 and FBMS59 correspondingly. We choose those samples from different videos as many as possible.
In Figure~\ref{fig:suppsaliency}, more saliency detection results from DUTS~\cite{wang2017learning} dataset are represented.

\subsection{Broader Impact}
We proposed a self-supervised pretraining method for zero-shot object segmentation. The central idea of decomposing appearance and motion can be implemented with other network architectures, and even training losses.
However, we have not studied the implications of these variations of the approach.
There will also be unpredictable failures,  where the generalization of the self-supervised framework still needs deeper understanding.
This method is data-driven thus the data bias problem should be careful during data collection in both pretraining and downstream tasks. As this method can be applied to a wide range of videos without annotation, privacy should be also careful during the data utilization. 

\vspace{1cm}
\begin{table}[H]
    \centering
    \caption{Details about the prediction head in our segmentation network. Our segmentation network consists of a backbone, ResNet50, and a prediction head which predicts the segments through the features from the backbone. Here $c$ is a hyperparameter which represents the segment number. }
    \vspace{4pt}
    \begin{tabular}{|c|c|}
    \hline
    Layer	&	Output size	\\
    \hline
    Input Feature	&	$2048 \times 48 \times 48$	\\
    Conv($3 \times 3, 2048 \rightarrow 256$) + BN + ReLU	&	$256\times 48 \times 48$	\\
    Conv($3 \times 3, 256 \rightarrow 256$) + BN + ReLU	&	$256 \times 48 \times 48$	\\
    Conv($3 \times 3, 256 \rightarrow c$)	&	$c\times 48 \times 48$	\\
    \hline
    \end{tabular}
    \label{tab:detailseg}
\end{table}

\begin{table}[t]
    \centering
    \caption{Architecture details about our correspondence network. As it processes the input at different pyramid levels, here $H$ and $W$ represents the size of input in a certain level. And $c$ is a hyperparameter about the segment number. }
    \vspace{4pt}
    \begin{tabular}{|c|c|c|}
    \hline
Index	&	Layer	&	Output size	\\
\hline
1.	&	Input Feature	&	$	115 \times H \times W	$	\\
2.	&	Conv($3 \times 3, 115 \rightarrow 128$) + ReLU	&	$	128 \times H \times W	$	\\
3.	&	Conv($3 \times 3, 128 \rightarrow 128$) + ReLU	&	$	128 \times H \times W	$	\\
4.	&	Concatenate 2. and 3.	&	$	256 \times H \times W	$	\\
5.	&	Conv($3 \times 3, 256 \rightarrow 96$) + ReLU	&	$	96 \times H \times W	$	\\
6.	&	Concatenate 3. and 5.	&	$	224 \times H \times W	$	\\
7.	&	Conv($3 \times 3, 224 \rightarrow 64$) + ReLU	&	$	64 \times H \times W	$	\\
8.	&	Concatenate 5. and 7.	&	$	160 \times H \times W	$	\\
9.	&	Conv($3 \times 3, 160 \rightarrow 32$) + ReLU	&	$	32 \times H \times W	$	\\
10.	&	Concatenate 7. and 9.	&	$	96 \times H \times W	$	\\
11.	&	Average Pooling	&	$	96 \times c	$	\\
12.	&	FC ($96 \rightarrow 2$)	&	$	2 \times c	$	\\
    \hline
    \end{tabular}
    
    \label{tab:detailpwc}
\end{table}

\def\rim#1#2#3#4#5#6#7#8{
\imw{figs_supp/#7/#8/#1.jpg}{0.152}&
\hspace{0.01cm}
\imw{figs_supp/#7/#8/#2.jpg}{0.152}&
\hspace{0.01cm}
\imw{figs_supp/#7/#8/#3.jpg}{0.152}&
\hspace{0.01cm}
\imw{figs_supp/#7/#8/#4.jpg}{0.152}&
\hspace{0.01cm}
\imw{figs_supp/#7/#8/#5.jpg}{0.152}&
\hspace{0.01cm}
\imw{figs_supp/#7/#8/#6.jpg}{0.152}\\
}
\begin{figure}[t]
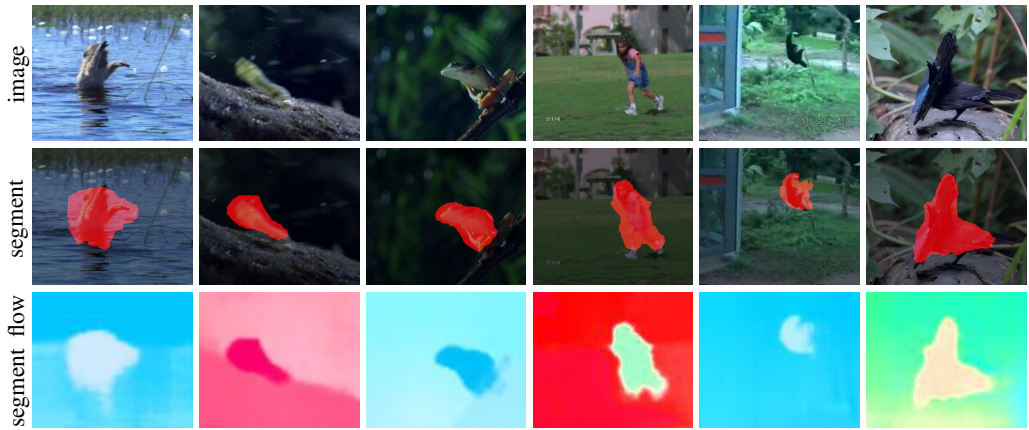

\centering
\subfloat{\small
    \tb{@{}ccccccc@{}}{0.03}{
        \rotatebox{90}{\hspace{0.5cm}image} & \rim{1}{4}{7}{10}{13}{16}{supp_segt}{im}
        \rotatebox{90}{\hspace{0.35cm}segment} & \rim{2}{5}{8}{11}{14}{17}{supp_segt}{mask}
        \rotatebox{90}{\makecell[c]{segment\hspace{0.15cm}flow}} & \rim{0}{3}{6}{9}{12}{15}{supp_segt}{f1}
    }
}
\caption{Qualitative results of SegTrackv2}
\label{fig:suppsegt}
\end{figure}

\def\rim#1#2#3#4#5#6#7{
\imw{figs_supp/#7/im/#1_iter0000100_im.jpg}{0.152}&
\hspace{0.01cm}
\imw{figs_supp/#7/im/#2_iter0000100_im.jpg}{0.152}&
\hspace{0.01cm}
\imw{figs_supp/#7/im/#3_iter0000100_im.jpg}{0.152}&
\hspace{0.01cm}
\imw{figs_supp/#7/im/#4_iter0000100_im.jpg}{0.152}&
\hspace{0.01cm}
\imw{figs_supp/#7/im/#5_iter0000100_im.jpg}{0.152}&
\hspace{0.01cm}
\imw{figs_supp/#7/im/#6_iter0000100_im.jpg}{0.152}\\
}
\def\rfl#1#2#3#4#5#6#7{
\imw{figs_supp/#7/f1/#1_iter0000100_f1.jpg}{0.152}&
\hspace{0.01cm}
\imw{figs_supp/#7/f1/#2_iter0000100_f1.jpg}{0.152}&
\hspace{0.01cm}
\imw{figs_supp/#7/f1/#3_iter0000100_f1.jpg}{0.152}&
\hspace{0.01cm}
\imw{figs_supp/#7/f1/#4_iter0000100_f1.jpg}{0.152}&
\hspace{0.01cm}
\imw{figs_supp/#7/f1/#5_iter0000100_f1.jpg}{0.152}&
\hspace{0.01cm}
\imw{figs_supp/#7/f1/#6_iter0000100_f1.jpg}{0.152}\\
}
\def\rmask#1#2#3#4#5#6#7{
\imw{figs_supp/#7/mask/#1_iter0000100_im_msk.jpg}{0.152}&
\hspace{0.01cm}
\imw{figs_supp/#7/mask/#2_iter0000100_im_msk.jpg}{0.152}&
\hspace{0.01cm}
\imw{figs_supp/#7/mask/#3_iter0000100_im_msk.jpg}{0.152}&
\hspace{0.01cm}
\imw{figs_supp/#7/mask/#4_iter0000100_im_msk.jpg}{0.152}&
\hspace{0.01cm}
\imw{figs_supp/#7/mask/#5_iter0000100_im_msk.jpg}{0.152}&
\hspace{0.01cm}
\imw{figs_supp/#7/mask/#6_iter0000100_im_msk.jpg}{0.152}\\
}
\begin{figure}[t]
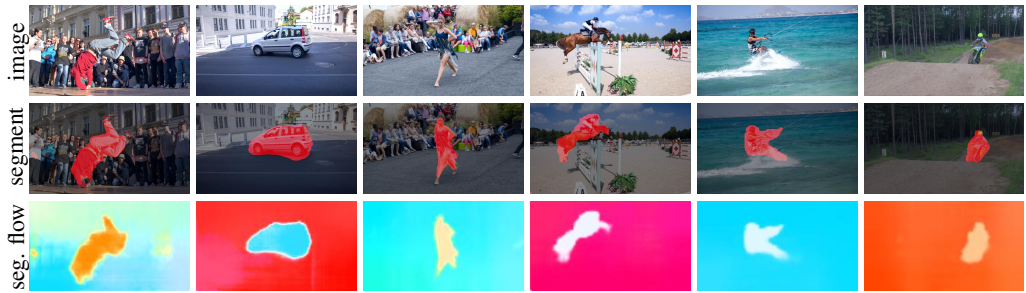

\centering
\subfloat{\small
    \tb{@{}ccccccc@{}}{0.03}{
        \rotatebox{90}{\hspace{0.2cm}image} & \rim{breakdance_00065}{car-shadow_00003}{dance-twirl_00000}{horsejump-high_00013}{kite-surf_00012}{motocross-jump_00000}{supp_davis}
        \rotatebox{90}{\hspace{0.1cm}segment} &  \rmask{breakdance_00065}{car-shadow_00003}{dance-twirl_00000}{horsejump-high_00013}{kite-surf_00012}{motocross-jump_00000}{supp_davis}
        \rotatebox{90}{\hspace{0.0cm}seg. flow} & \rfl{breakdance_00065}{car-shadow_00003}{dance-twirl_00000}{horsejump-high_00013}{kite-surf_00012}{motocross-jump_00000}{supp_davis}
    }
}
\caption{Qualitative results of DAVIS 2016}
\label{fig:suppdavis}
\end{figure}

\def\rim#1#2#3#4#5#6#7#8{
\imw{figs_supp/#7/#8/#1.jpg}{0.152}&
\hspace{0.01cm}
\imw{figs_supp/#7/#8/#2.jpg}{0.152}&
\hspace{0.01cm}
\imw{figs_supp/#7/#8/#3.jpg}{0.152}&
\hspace{0.01cm}
\imw{figs_supp/#7/#8/#4.jpg}{0.152}&
\hspace{0.01cm}
\imw{figs_supp/#7/#8/#5.jpg}{0.152}&
\hspace{0.01cm}
\imw{figs_supp/#7/#8/#6.jpg}{0.152}\\
}
\begin{figure}[t]
\centering
\subfloat{\small
    \tb{@{}ccccccc@{}}{0.03}{
        \rotatebox{90}{\hspace{0.5cm}image} & \rim{1}{4}{7}{10}{13}{16}{supp_fbms}{im}
        \rotatebox{90}{\hspace{0.35cm}segment} & \rim{2}{5}{8}{11}{14}{17}{supp_fbms}{mask}
        \vspace{0.3cm}
        \rotatebox{90}{\makecell[c]{segment\hspace{0.15cm}flow}} & \rim{0}{3}{6}{9}{12}{15}{supp_fbms}{f1}
        \rotatebox{90}{\hspace{0.5cm}image} & \rim{19}{22}{25}{28}{31}{34}{supp_fbms}{im}
        \rotatebox{90}{\hspace{0.35cm}segment} & \rim{20}{23}{26}{29}{32}{35}{supp_fbms}{mask}
        \rotatebox{90}{\makecell[c]{segment\hspace{0.15cm}flow}} & \rim{18}{21}{24}{27}{30}{33}{supp_fbms}{f1}
    }
}
\caption{Qualitative results of FBMS59}
\label{fig:suppfbms}
\end{figure}

\def\prow#1#2#3#4{
\imw{figs_supp/supp_saliency/#1.png}{0.25}&
\hspace{0.01cm}
\imw{figs_supp/supp_saliency/#2.png}{0.25}&
\hspace{0.01cm}
\imw{figs_supp/supp_saliency/#3.png}{0.25}&
\hspace{0.01cm}
\imw{figs_supp/supp_saliency/#4.png}{0.25}\\
}
\begin{figure}[t]
\centering
\subfloat{\small
    \tb{@{}cccc@{}}{0.05}{
    \prow{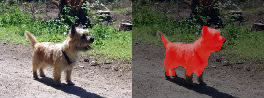}{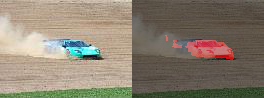}{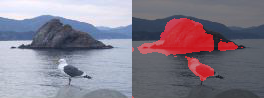}{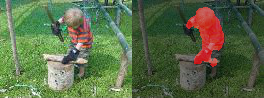}
    \prow{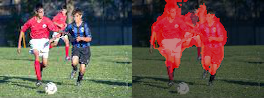}{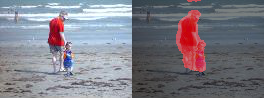}{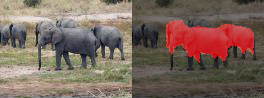}{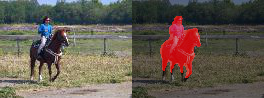}
    \prow{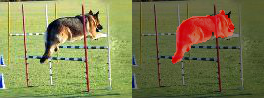}{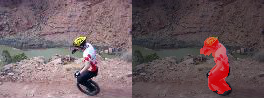}{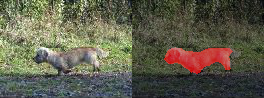}{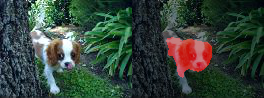}
    \prow{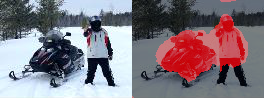}{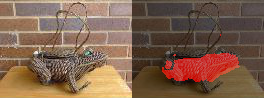}{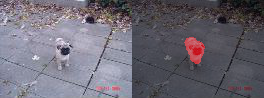}{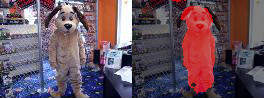}
    \prow{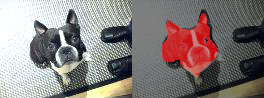}{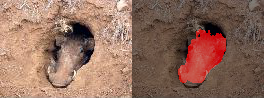}{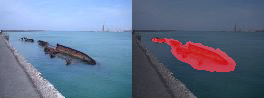}{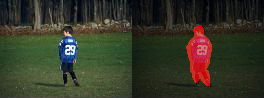}
    \prow{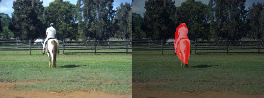}{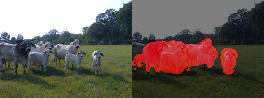}{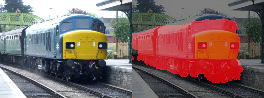}{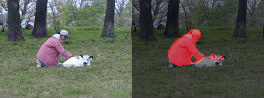}
    }
}
\caption{Qualitative salient object detection results. Our model can detect multiple primary objects and even static object like the chair and the rocks. }
\label{fig:suppsaliency}
\end{figure}

\end{document}